\documentclass[conference]{IEEEtran}
\IEEEoverridecommandlockouts
\usepackage{amsmath,amssymb,amsfonts}
\usepackage{algorithmic}
\usepackage{graphicx}
\usepackage{textcomp}
\usepackage{tabularx}
\usepackage{float}
\usepackage{booktabs}
\usepackage[backend=bibtex,sorting=none]{biblatex}
\usepackage[table]{xcolor}  

\usepackage{amssymb} 

\def\BibTeX{{\rm B\kern-.05em{\sc i\kern-.025em b}\kern-.08em
    T\kern-.1667em\lower.7ex\hbox{E}\kern-.125emX}}

\begin{document}

\title{EmoAssist: Emotional Assistant for Visual Impairment Community}

\author{
\IEEEauthorblockN{Xingyu Qi\textsuperscript{1, 2}, He Li\textsuperscript{1, 2}, Linjie Li\textsuperscript{1} and Zhenyu Wu\textsuperscript{1}}
\IEEEauthorblockA{
\textit{\textsuperscript{1}Beijing University of Posts and Telecommunications} \\
\textit{\textsuperscript{2}AI2Robotics} \\
\{qixingyu, lihe2024, lilinjie, shower0512\}@bupt.edu.cn}
}


\maketitle

\begin{abstract}
The rapid advancement of large multi-modality models (LMMs) has significantly propelled the integration of artificial intelligence into practical applications. Visual Question Answering (VQA) systems, which can process multi-modal data including vision, text, and audio, hold great potential for assisting the Visual Impairment (VI) community in navigating complex and dynamic real-world environments. However, existing VI assistive LMMs overlook the emotional needs of VI individuals, and current benchmarks lack emotional evaluation of these LMMs. To address these gaps, this paper introduces the EmoAssist Benchmark, a comprehensive benchmark designed to evaluate the assistive performance of LMMs for the VI community. To the best of our knowledge, this is the first benchmark that incorporates emotional intelligence as a key consideration. Furthermore, we propose the EmoAssist Model, an Emotion-Assistive LMM specifically designed for the VI community. The EmoAssist Model utilizes Direct Preference Optimization (DPO) to align outputs with human emotional preferences. Experiment results demonstrate that the EmoAssist Model significantly enhances the recognition of implicit emotions and intentions of VI users, delivers empathetic responses, and provides actionable guidance. Specifically, it shows respective improvements of 147.8\% and 89.7\% in the Empathy and Suggestion metrics on the EmoAssist Benchmark, compared to the pre-tuning LMM, and even outperforms state-of-the-art LLMs such as GPT-4o.

\end{abstract}


\section{Introduction}
Visual impairment (VI) refers to the partial or total inability of visual perception \cite{1}. According to the International Agency for the Prevention of Blindness \cite{2}, by 2050, the number of VI individuals will reach 1.758 billion, including 61 million with blindness and 474 million with moderate to severe VI. With the advent of the artificial intelligence era, many efforts have been made to develop assistive models and systems for the VI community \cite{3,4,5,9,10,14,16,22}. However, existing works still have certain limitations, and this paper continues to focus on assistive LMMs for the VI community.

\begin{figure}
    \centering
    \includegraphics[width=1\linewidth]{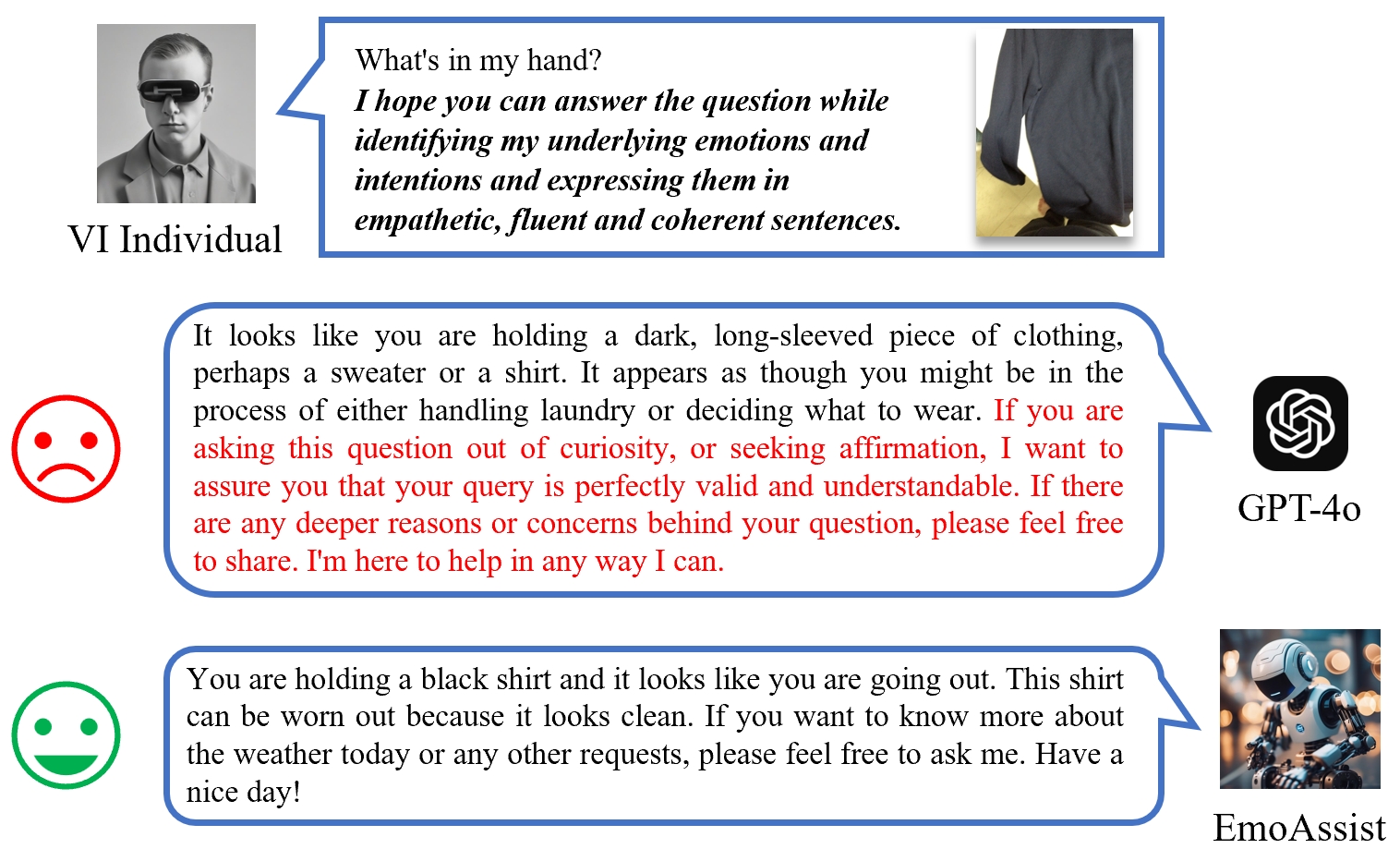}
    \caption{Emotional intelligence performance of GPT-4o and the EmoAssist Model on VI assistance}
    \label{fig:EmoAssist_vs_GPT-4o}
\end{figure}

From the perspective of evaluating VI assistive LMMs, existing work \cite{18,19,20} is mainly based on two evaluation approaches, which are automatic evaluation and human evaluation. Automatic evaluation typically employs metrics such as ROUGE \cite{36}, METEOR \cite{37}, and BERT Score \cite{38}, which measure token overlap or semantic similarity between generated and reference texts. Recent studies such as LAVE \cite{35}, have also utilized GPT-4 \cite{15} as a scorer. Human evaluation involves subjective assessments based on criteria such as correctness, fluency, and actionability. However, none of these existing evaluation frameworks incorporate emotional needs into the benchmark. As a vulnerable disabled group, VI individuals face significant psychological challenges, and existing research \cite{24,25} has underscored their urgent need for emotional support and empathetic responses in VQA \cite{8} scenarios. Addressing this gap, we propose the \textbf{EmoAssist Benchmark}, a comprehensive framework to evaluate the assistive performance of LMMs. Notably, we are the first to integrate emotional intelligence into the evaluation criteria. Our benchmark identifies five key dimensions—Relevance, Empathy, Suggestion, Coherence, and Fluency, derived from the real-world needs and experiences of VI individuals. To facilitate a robust evaluation process, we have constructed a high-quality dataset named the \textbf{EmoAssist Dataset} containing 1,000 samples through a combination of GPT-based augmentation and human review. By incorporating automatic evaluation and human scoring, our benchmark provides reliable and quantitative results to assess the assistive performance of LMMs.

\begin{figure}
    \centering 
    \includegraphics[width=0.9\linewidth]{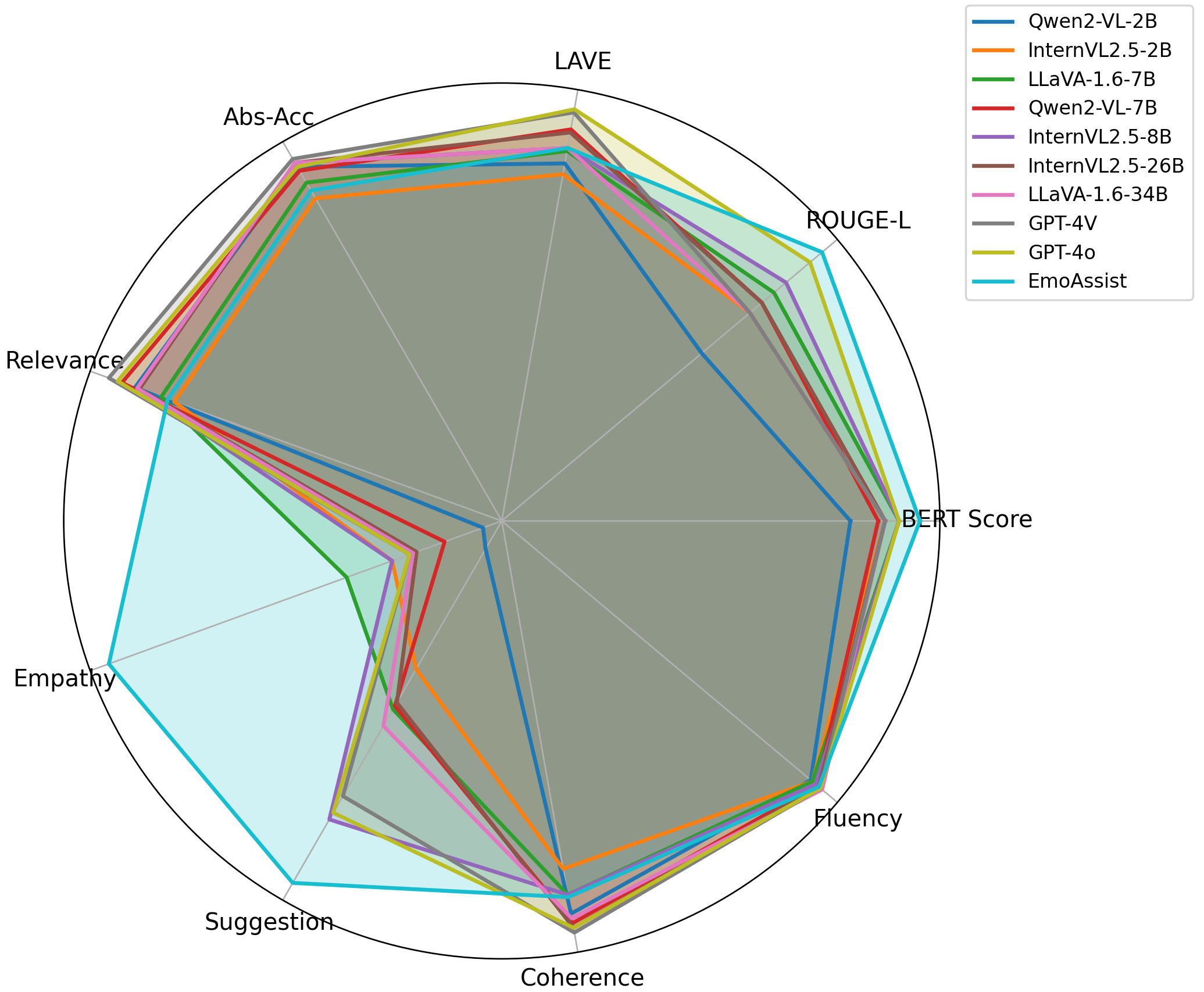}
    \caption{LMMs Performance on the EmoAssist Benchmark}
    \label{fig:total}
\end{figure}

From the perspective of emotional intelligence in VI assistive LMMs, as shown in Figure \ref{fig:EmoAssist_vs_GPT-4o}, even GPT-4o \cite{6}, a state-of-the-art LMM, struggles to accurately comprehend the intentions of VI users and respond with empathy, resulting in confusing and inadequate responses. Furthermore, Figure \ref{fig:total} presents the evaluation results of various LMMs on the EmoAssist Benchmark, highlighting their poor performance in terms of Empathy and Suggestion. Specifically, these models tend to focus primarily on delivering accurate descriptions of visual content, but fall short in recognizing users' implicit emotional needs and providing empathetic responses. They also struggle to interpret the underlying intentions of VI individuals, thus failing to offer appropriate guidance. Motivated by these limitations, we propose the \textbf{EmoAssist Model}, a high-performance assistive LMM specifically designed for VQA scenarios involving VI users. The EmoAssist Model excels not only in delivering accurate and fluent information but also in providing highly empathetic responses, effectively addressing the emotional needs of VI individuals. To achieve this, we leverage the Low-Rank Adaptation (LoRA) \cite{33} technique for parameter-efficient fine-tuning and apply Direct Preference Optimization (DPO) \cite{34} to train LLaVA \cite{7} to better understand and prioritize human emotional preferences.

Experimental results demonstrate the effectiveness of our training approach and the outstanding performance of the EmoAssist Model. The EmoAssist Model achieved scores of 1.14 and 1.10 in Empathy and Suggestion metrics of the EmoAssist Benchmark respectively, showing improvements of 147.8\% and 89.7\% compared to the pre-tuning LMM. Additionally, it outperforms GPT-4o, the state-of-the-art model, with increases of +0.68 in Empathy and +0.19 in Suggestion.

To the best of our knowledge, our work is the first to incorporate emotional intelligence into LMMs for assistive technologies designed for the VI community, the key contributions of our paper are as follows.
\begin{itemize}
    \item We propose the EmoAssist Benchmark, a comprehensive benchmark that is the first to incorporate the emotional needs of VI individuals in evaluating the assistive capabilities of LMMs, while also integrating prior evaluations grounded in real-world scenarios.
    \item To address the challenge of limited emotional intelligence in existing LMMs, we introduce the EmoAssist Model, an assistive LMM fine-tuned using LoRA and DPO, designed to exhibit high emotional intelligence.
    \item Experimental results demonstrate the EmoAssist Model's effectiveness in enhancing the VI individuals' experience by understanding emotions and intents and providing empathetic responses and accessible suggestions.

\end{itemize}

\section{Related Work}

\subsection{Visual Question Answering (VQA)}
In recent years, the exceptional capabilities of LMMs such as GPT-4o and LLaVA have garnered significant attention due to their remarkable visual understanding, reasoning abilities, and natural interaction styles. VQA has emerged as a promising technology, which generates accurate and meaningful answers based on visual input and natural language questions posed by users. This form of interaction is very suitable for assisting VI individuals. For instance, a VI individual can take a picture of their surroundings and ask questions such as \textit{"What obstacles are here?"} or \textit{"Is the traffic light ahead red or green?"} The VQA system powered by LMMs can respond to these questions by generating answers based on the visual content, thus enabling the VI individuals to better understand and assess their environment in real time. With the the development of VQA techniques, several VQA datasets have contributed, such as DAQUAR \cite{9}, VQAv2 \cite{10}, FM-IQA \cite{11}, CLEVR \cite{12}, VizWiz-VQA \cite{13}.

\subsection{Evaluating Emotional Intelligence }
\textit{Wang et al.} \cite{40} proposed a novel psychometric evaluation method focusing on emotional understanding and tested it on various large language models (LLMs). The results showed that most LLMs outperformed the average level of human emotional intelligence, with GPT-4 scoring higher than 89\% of humans. \textit{Sabour et al.} introduced EmoBench \cite{41}, a benchmark dataset containing more than 400 meticulously crafted questions aimed at assessing LLMs’ capabilities in emotional reasoning and understanding. Furthermore, \textit{Chen et al.} proposed EmotionQueen \cite{26}, which categorizes emotional intelligence assessment into four subtasks, which are key event recognition, mixed event recognition, implicit emotional recognition, and intention recognition. They also designed two evaluation metrics to measure LLM's performance in recognizing and responding to emotion-related statements.

\subsection{AI for VI Assistance}
Before the advent of large models (LMs), various AI-based systems were developed to assist the VI community. Notable examples include DeepNAVI \cite{3}, V-eye \cite{4}, and the VI assistive system proposed by \textit{Lin et al.} \cite{5}, all of which aimed to provide scene understanding and navigation support for VI individuals. With the rise of LMs, OpenSU \cite{22} integrated SAM \cite{23} to offer more comprehensive scene descriptions, significantly enhancing the independent mobility of VI individuals. Be My AI \cite{14} and VQAsk \cite{16} leverage LMMs not only to describe images but also to perform VQA tasks for VI users. The "Hear World" app , powered by MouSi \cite{17}, offers a wider range of functionalities tailored for the VI community, including VQA, item search, and text recognition.

\section{EmoAssist Benchmark: Comprehensive Benchmark of LMM's Assistance Capabilities for the VI Community}

\begin{figure}[h!]
    \centering
    \includegraphics[width=0.95\linewidth]{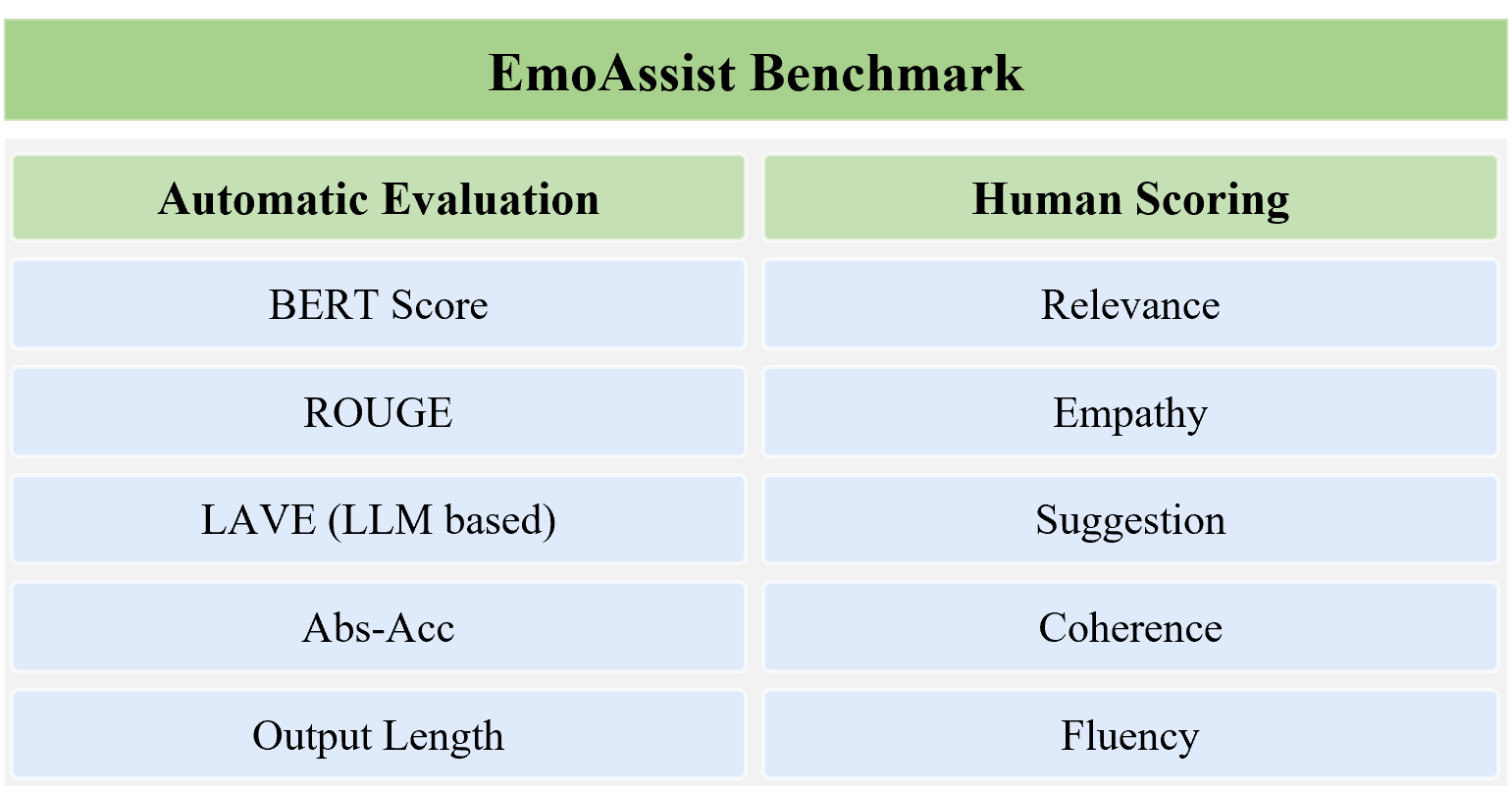}
    \caption{EmoAssist Benchmark}
    \label{fig:bench}
\end{figure}

To comprehensively evaluate the ability of LMM to assist VI individuals, especially by adding emotional intelligence evaluation indicators on the basis of prior work, we proposed the EmoAssist Benchmark, which is divided into automatic evaluation and human scoring, as shown in Figure \ref{fig:bench}. Additionally, we constructed the EmoAssist Dataset to perform evaluation and further tuning of the EmoAssist Model.

\subsection{Automatic Evaluation}

Following \textit{Mina Huh et al.} \cite{20}, we adopt \textbf{BERT Score}, \textbf{ROUGE}, \textbf{LAVE}, Abstain Accuracy (\textbf{Abs-Acc}) and \textbf{Output Length} as automatic evaluation metrics.

BERT Score leverages contextual embeddings of BERT to measure the semantic similarity between candidates and reference texts. In the following equation, \( P \) represents the set of BERT embeddings for the generated text, and  
\( Q \) represents the set of embeddings for the reference text,  
\( \text{sim}(p, q) \) denotes the cosine similarity between two vectors,  
and \( \max_{q \in Q} \) indicates that, for each generated text embedding \( p \), the most similar reference text embedding \( q \) is selected.

\begin{equation}
\text{BERTScore}(P, Q) = \frac{1}{|P|} \sum_{p \in P} \max_{q \in Q} \text{sim}(p, q)
\end{equation}

ROUGE is a widely used evaluation metric in natural language generation tasks, designed to measure the degree of overlap between texts. In the formula for ROUGE-N, \( N \) refers to the n-gram size, \( n \) and \( n' \) are n-grams from the generated and reference text, respectively, \( \text{count}(n' \in \text{generated}) \) counts how often the n-gram \( n' \) appears in the generated text.

\begin{equation} 
  \text{ROUGE-N} = \frac{\sum_{n \in \text{generated}} \sum_{n' \in \text{reference}} \text{count}(n' \in \text{generated})}{\sum_{n \in \text{generated}} \text{count}(n)} 
\end{equation}

LAVE is an LLM-based metric that utilizes GPT-4 to evaluate a candidate answer against a reference answer. Since VI individuals often capture images that are unanswerable, we include Abs-Acc as an automatic evaluation metric, which measures whether the abstain decisions given by LMM are correct against the gold answerability label.

\subsection{Human Scoring}

Existing studies \cite{18,20} have evaluated LMMs through human evaluations, considering aspects such as correctness, actionability, fluency, clarity, relevance, helpfulness and plausibility. In these evaluations, human annotators typically score the outputs or rank multiple LMMs to compare their performance.

However, a comprehensive analysis of these studies reveals a lack of incorporation of emotional intelligence into their evaluation frameworks, with some metrics demonstrating significant overlap. Building upon this analysis, we propose five key dimensions as human scoring criteria for the EmoAssist Benchmark: Relevance, Empathy, Suggestion, Coherence, and Fluency. Among these, Empathy and Suggestion are critical indicators reflecting the emotional intelligence of assistive LMMs, inspired by EmoBench \cite{41} and EmotionQueen \cite{26}. To evaluate the responses generated by LMMs, we employ human annotators to score based on these five criteria, using a scale ranging from 0 to 2, detailed scoring guidelines and annotations are shown in Table \ref{tab:five key}. 

\begin{enumerate}
    \item \textbf{Relevance} \cite{20,27,28}: Assesses whether the LMM's responses address the question directly while balancing emotional expression with contextual relevance.
    
    \item \textbf{Empathy} \cite{29}: Measures the ability to understand VIs' emotions and respond with warmth, compassion, and concern, addressing their unique challenges.
    
    \item \textbf{Suggestion} \cite{18,20,26}: Evaluates the capacity to understand the intent behind VIs' queries and provide actionable, specific, and tailored suggestions.
    
    \item \textbf{Coherence} \cite{30,31}: Reflects the logical flow and clarity of responses, ensuring they are free from repetition or contradictions, crucial for meaningful dialogue.
    
    \item \textbf{Fluency} \cite{20,30,32}: Assesses grammatical correctness, structure, and readability, ensuring smooth and effective communication.
\end{enumerate}

\begin{table*}[h]
\caption{Human scoring metrics of the EmoAssist Benchmark}
\label{tab:five key}
\centering
\renewcommand{\arraystretch}{1.5} 
\setlength{\tabcolsep}{2pt} 
\begin{tabularx}{\textwidth}{|>{\centering\arraybackslash}m{2.5cm}|>{\centering\arraybackslash}m{3.77cm}|>{\centering\arraybackslash}m{3.7cm}|>{\centering\arraybackslash}m{3.7cm}|>{\centering\arraybackslash}m{3.7cm}|}
  \hline
  \textbf{Dimension} & \textbf{Explanation} & \textbf{0 Score} & \textbf{1 Score} & \textbf{2 Score} \\
  \hline
  \textbf{Relevance} 
  & Measures how relevant the answer is to the question.  
  & The answer has nothing to do with the question.  
  & Only part of the answer is related to the question. 
  & The answer as a whole is relevant to the question. \\
  \hline
  \textbf{Empathy} 
  & Ability to understand users' feelings or implicit emotion and respond in warmth, compassion, and concern.  
  & The answer does not demonstrate any understanding of the user's feelings or hidden emotions and the response is not warm. 
  & The answer demonstrates that it tries to understand the user's feelings, but the understanding is faulty or incomplete. The answer is not considerate or warm enough.  
  & The answer fully understands users' feelings or implicit emotions and responds in warmth, compassion, and concern. \\
  \hline
  \textbf{Suggestion} 
  & Ability to understand the intention of the user and give suggestions which are available for VI.  
  & The answer does not understand the user's intent and does not give any advice.  
  & The answer doesn't fully understand the user's intent and gives some simple suggestions. But those suggestions are not reasonable or practical for the blind person. 
  & The answer fully understands the user's intent and gives suggestions, which are helpful and actionable for the blind person. \\
  \hline
  \textbf{Coherence} 
  & Coherence refers to the degree to which a response is highly coherent, meaning it has clear logic, no repetition or redundancy, and pronoun references are clear. 
  & The answer is extremely incoherent and riddled with logical errors that seriously detracted from the reading experience.  
  & The answer overall is coherent, although there were some logic issues that didn't detract from the reading. 
  & The answer is highly coherent with no logical problems. \\
  \hline
  \textbf{Fluency} 
  & Fluency refers to whether the response is easy to read. 
  & The answer is completely not fluent, with unnatural expressions, grammar mistakes, and improper word choices throughout, making it difficult to read. 
  & The answer is generally fluent. Although there are some grammar errors and inaccurate word choices, it is not difficult to read. 
  & The answer is completely fluent. There are no grammar errors, the word choices are accurate, and it is natural and quite easy to read. \\
  \hline
\end{tabularx}
\end{table*}

\begin{figure*}
    \centering
    \includegraphics[width=0.8\linewidth]{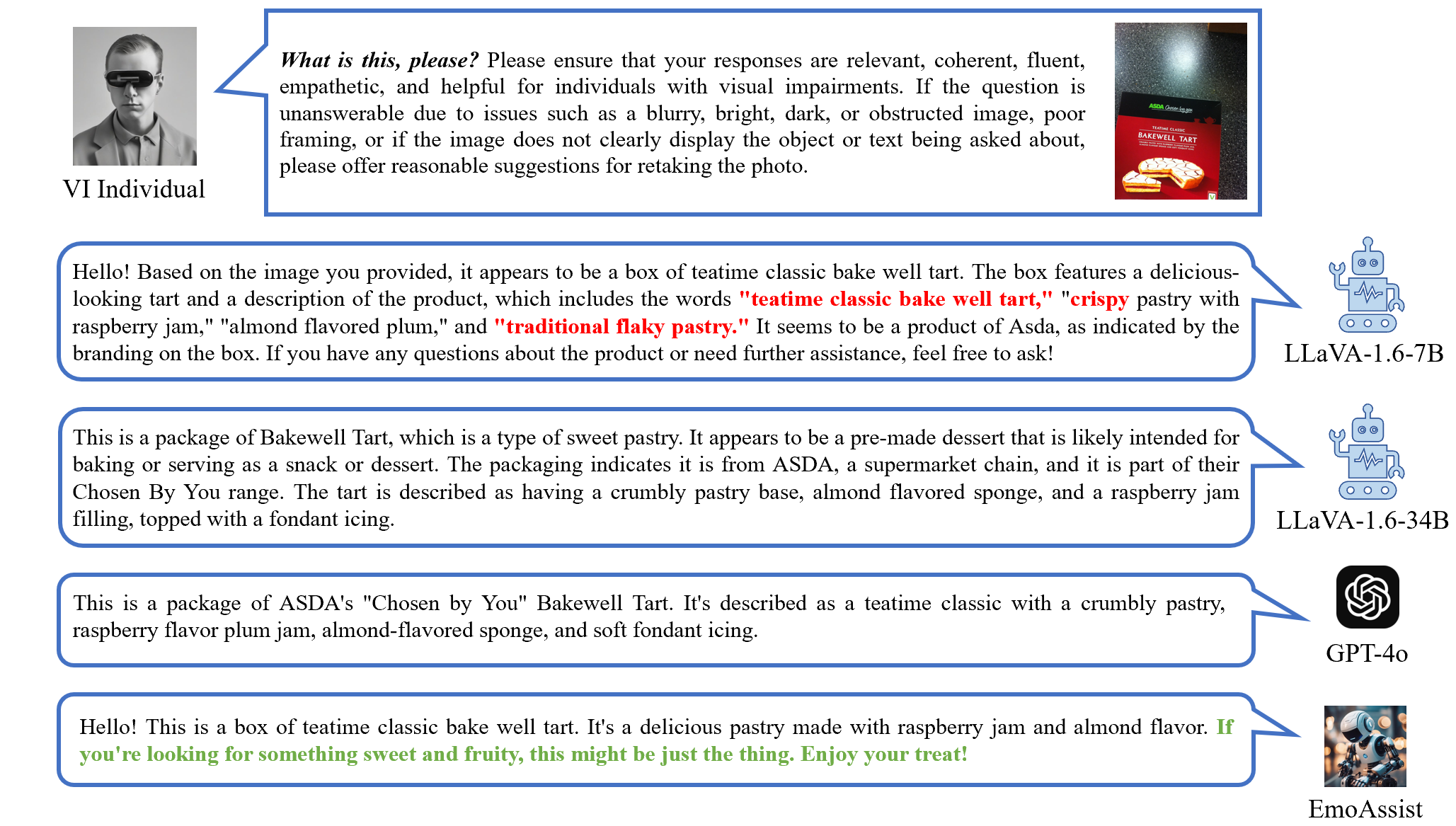}
    \caption{EmoAssist Model performance on VI individual queries compared with baseline LMMs}
    \label{fig:EmoAssist_vs_3_baselines}
\end{figure*}

\subsection{EmoAssist Dataset Construction}

To conduct the EmoAssist Benchmark evaluation and tuning of the EmoAssist Model, we construct the EmoAssist Dataset.

We selected the VizWiz-VQA dataset as our data source, as it is specifically designed for assistive technology in VI scenarios, with all photos taken by VI individuals. The dataset comprises 31,173 QA pairs, with an average of 6.7 words per question and 1.7 words per answer. However, the answers in VizWiz-VQA lack empathetic elements, making them unsuitable for direct use in our study. Therefore, we performed several data pre-processing and construction steps.

First, we filtered low-quality data from the VizWiz-VQA dataset. As each VizWiz-VQA answer consists of ten crowd-sourced responses, we excluded QA pairs where the majority vote was fewer than five. Additionally, we removed questions containing the conjunction \textit{"and"} with more than five words, as well as image-question pairs with incomplete queries, such as \textit{"Is this?"}. Duplicate questions, such as \textit{"What is this? What is this?"}, were consolidated into a single query \textit{"What is this?"}. Second, we retained unanswerable questions and capped their proportion at 30\% to maintain consistency with the original VizWiz-VQA dataset. Third, to better align with our emotional use case, we prompted GPT-4o to generate complete sentences from the extracted words or phrases, ensuring that the responses expressed emotions towards VI individuals. This was followed by a manual review to further enhance the dataset quality. Finally, the EmoAssist Dataset is formatted as \texttt{\textless Image, Question, Ground Truth\textgreater}, enabling its use in evaluating the performance of LMMs and further tuning of the EmoAssist Model.

\section{EmoAssist Model: Adapting LLaVA as Emotional Assistant}

\subsection{Preliminary: LoRA and DPO}

Low-Rank Adaptation (LoRA) introduces a method for efficiently fine-tuning large pretrained models by decomposing weight matrices into low-rank components. This technique reduces both computational costs and the number of parameters while maintaining performance. 

Direct Preference Optimization (DPO) optimizes models based on human preferences by directly learning from comparative feedback, rather than relying solely on traditional loss functions. This allows models to better align with user preferences in tasks like text generation, improving the relevance and quality of responses.

\begin{table*}[htbp]
    \caption{LMMs Proformance on the EmoAssist Benchmark}
    \centering
    \label{tab:result}
    \renewcommand{\arraystretch}{1.5} 
    \resizebox{\textwidth}{!}{ 
    \begin{tabular}{lcccccccccccc}
        \hline
        LMM & BERT Score & ROUGE-1 & ROUGE-2 & ROUGE-L & LAVE & Abs-Acc & Relevance & Empathy & Suggestion & Coherence & Fluency & Output Len \\
        \hline
        Qwen2-VL-2B & 0.51 & 0.23 & 0.08 & 0.17 & 2.35 & 0.91 & 1.71 & 0.07 & 0.10 & 1.69 & 1.93 & 25.68 \\
        InternVL2.5-2B & 0.56 & 0.31 & 0.10 & 0.21 & 2.28 & 0.83 & 1.53 & 0.33 & 0.46 & 1.50 & 1.94 & 65.89 \\
        LLaVA-1.6-7B & \underline{0.58} & 0.33 & 0.11 & 0.23 & 2.43 & 0.87 & 1.59 & \underline{0.46} & 0.58 & 1.61 & 1.94 & 67.92 \\
        Qwen2-VL-7B & 0.55 & 0.30 & 0.11 & 0.22 & 2.57 & 0.90 & 1.77 & 0.18 & 0.57 & 1.73 & 1.97 & 45.40 \\
        InternVL2.5-8B & \underline{0.58} & \underline{0.35} & \underline{0.12} & 0.24 & 2.45 & \underline{0.92} & 1.70 & 0.33 & \underline{0.91} & 1.61 & 1.96 & 56.10 \\
        InternVL2.5-26B & 0.56 & 0.31 & 0.11 & 0.22 & 2.55 & \underline{0.92} & 1.69 & 0.26 & 0.56 & \underline{1.75} & 1.98 & 45.28 \\
        LLaVA-1.6-34B & 0.56 & 0.32 & 0.10 & 0.21 & 2.45 & \underline{0.92} & 1.70 & 0.27 & 0.63 & 1.71 & \textbf{2.00} & 69.46 \\
        GPT-4V & 0.56 & 0.32 & 0.10 & 0.21 & \underline{2.68} & \textbf{0.93} & \textbf{1.83} & 0.28 & 0.84 & \textbf{1.77} & \underline{1.99} & 84.04 \\
        GPT-4o & \underline{0.58} & 0.34 & \textbf{0.15} & \underline{0.26} & \textbf{2.70} & 0.91 & \underline{1.79} & 0.28 & 0.89 & \underline{1.75} & \underline{1.99} & 40.02 \\
        \rowcolor{gray!30} 
        \textbf{EmoAssist-7B} & \textbf{0.61} & \textbf{0.39} & \textbf{0.15} & \textbf{0.27} & 2.45 & 0.85 & 1.56 & \textbf{1.14} & \textbf{1.10} & 1.62 & 1.98 & 56.98 \\
        \hline
    \end{tabular}
    }
\end{table*}

\subsection{Fine-tuning LMM with LoRA and DPO}

Given the large number of trainable parameters in LMM, full-parameter fine-tuning is costly. Therefore, we chose to use LoRA for fine-tuning to significantly reduces the number of trainable parameters and improves training efficiency.

To enhance the LMM's performance, DPO algorithm was implemented. During the DPO phase, each training instance was structured in the format \texttt{\textless Image, Question, Chosen Response, Rejected Response\textgreater}, and fine-tuning objectives were two-fold: first, to maximize the alignment of the model's output preference with the chosen response while minimizing alignment with the rejected response. Concurrently, the DPO algorithm employed a reference model, initialized as a duplicate of the original model, and introduced a regularization term in the parameter update process. This regularization ensured that the model retained its inherent linguistic capabilities and domain knowledge while adapting to human preference signals.

In the context of VQA task targeting at VI community, we believe that the accuracy of existing LMMs is sufficient to meet requirements, given their demonstrated capabilities in recent advancements. Our fine-tuning objective is to enhance the LMM's emotional intelligence and empathetic response capabilities. DPO algorithm is particularly well-suited for this purpose. Specifically, we use the output of the untuned LMM as the rejected response, given its suboptimal performance on two emotional intelligence metrics—Empathy and Suggestion (as illustrated in Figure \ref{fig:total}). In contrast, the Ground Truth responses from the EmoAssist Dataset are employed as the chosen responses. This design is predicated on the observation that the chosen responses demonstrate significantly higher emotional intelligence than the original LMM outputs. By leveraging contrastive style tuning, the LMM is better equipped to generate responses with enhanced emotional intelligence and empathy.

\section{Experiment}

\subsection{EmoAssist Dataset Statistics}

We randomly selected 1,000 samples from the EmoAssist Dataset, comprising 700 answerable and 300 unanswerable pairs. 200 samples (140 answerable and 60 unanswerable) were reserved for comprehensive LMM evaluations on the EmoAssist Benchmark, while the remaining 800 samples (560 answerable and 240 unanswerable) were allocated for LoRA and DPO tuning of the EmoAssist Model.

\subsection{Implementation Details}

The EmoAssist Model is based on the LLaVA-1.6-Vicuna-7B, which integrates Vicuna-7B (an LLM) with a pretrained visual encoder, CLIP ViT-L/14 with a two-layer MLP and LoRA module. The MLP layer serves as a trainable projection matrix, mapping visual features from the CLIP encoder into the word embedding space of the Vicuna-7B.

The EmoAssist Model was tuned on 800 samples using 8 A100 GPUs with a batch size of 32, a learning rate of 5e-6, the AdamW optimizer, and a DPO \( \beta \) setting of 0.1. The training process was completed in 2 hours. The entire process aimed to empower the EmoAssist Model with the ability to recognize users' implicit emotions and intentions and generate empathetic responses and high quality guidance. Additionally, it sought to enhance its overall VQA capabilities, such as accurately identifying the number of items, recognizing text, providing reasonable reshooting suggestions when faced with unanswerable questions.

\subsection{Evaluation Metric and Baselines}

We used the EmoAssist Benchmark as the evaluation metric, as it comprehensively measures the assistive capabilities of LMMs for the VI community.

We selected currently popular open-source and closed-source LMMs as comparison models, including LLaVA-1.6, InternVL2.5 \cite{42}, Qwen2-VL \cite{43}, and the GPT series LMMs. For open source models, we also conducted experiments with different parameter sizes (ranging from 2B to 34B). We selected LLaVA-1.6-7B, LLaVA-1.6-34B, and GPT-4o as baseline LMMs, as the EmoAssist Model is fine-tuned on top of LLaVA, and GPT-4o is currently recognized as the state-of-the-art LMM.

\subsection{Performance on VI Assistive VQA}

\subsubsection{Qualitative Results}

Figure \ref{fig:EmoAssist_vs_3_baselines} presents the performance of EmoAssist compared with three baseline LMMs. The results demonstrate that the EmoAssist Model's responses exhibit two distinct advantages. First, the EmoAssist Model provides accurate and concise answers, whereas LLaVA-7B introduces hallucinated content along with excessive redundant information. Second, the EmoAssist Model demonstrates a high degree of emotional intelligence by effectively recognizing both the user's implicit emotions and intentions—for example, identifying the user’s desire for \textit{"something sweet"}—and responding with empathy and warmth. In contrast, LLaVA-34B and GPT4-o, while capable of accurately conveying information, fail to exhibit the same level of emotional intelligence as the EmoAssist Model.

\subsubsection{Quantitative Results}

As presented in Table \ref{tab:result}, the EmoAssist Model outperforms all comparison LMMs across multiple automatic evaluation metrics, including BERT Score, ROUGE-1, ROUGE-2, and ROUGE-L. Specifically, the EmoAssist Model achieves a BERT Score of 0.61, which is 0.03 higher than the best-performing LMM. It also attains a ROUGE-1 score of 0.39, representing a 0.03 improvement over the top-performing baseline model. These results highlight that, through fine-tuning, the EmoAssist Model is capable of generating responses for VI individuals that are not only more precise but also contextually enriched, ensuring alignment with their specific needs and preferences.

In terms of key emotional intelligence indicators, namely Empathy and Suggestion, the EmoAssist Model demonstrates exceptional performance, achieving scores of 1.14 and 1.10, respectively. Compared to its base model, LLaVA-v1.6-7B, the EmoAssist Model achieves remarkable improvements of 147.8\% in Empathy and 89.7\% in Suggestion. Furthermore, when compared to the best-performing baseline model, the EmoAssist Model achieves additional improvements of +0.68 and +0.19 in these metrics respectively. These results underscore the EmoAssist Model's exceptional capacity to accurately recognize and respond to the emotional states and underlying intentions of users, particularly those with VI. Its responses not only demonstrate a profound understanding of the user's emotional context but also provide actionable, context-sensitive guidance delivered with empathy and compassion.

\section{Conclusion and Future Work}

This work presents the EmoAssist Benchmark, a comprehensive evaluation benchmark designed to assess the overall capabilities of assistive systems for the VI community, along with the EmoAssist Model, an LMM demonstrating high emotional intelligence, as evidenced by both qualitative and quantitative experimental results. To the best of our knowledge, this is the first study to integrate the emotional intelligence needs of the VI community into the domain of assistive LMM. Future efforts will focus on expanding EmoAssist to support additional modalities, such as auditory and tactile interfaces, as well as integrating it with smart devices to deliver practical, real-world solutions for VI users, thereby fostering a more inclusive and supportive environment.

\printbibliography

@misc{1,
    author="WHO",
    title="Blindness and vision impairment",
    year="2024",
    url="https://www.who.int/news-room/fact-sheets/detail/blindness-and-visual-impairment"
}

@article{2,
  title={Global prevalence of visual impairment associated with myopic macular degeneration and temporal trends from 2000 through 2050: systematic review, meta-analysis and modelling},
  author={Fricke, Timothy R and Jong, Monica and Naidoo, Kovin S and Sankaridurg, Padmaja and Naduvilath, Thomas J and Ho, Suit May and Wong, Tien Yin and Resnikoff, Serge},
  journal={British Journal of Ophthalmology},
  volume={102},
  number={7},
  pages={855--862},
  year={2018},
  publisher={BMJ Publishing Group Ltd}
}

@article{3,
title = {DeepNAVI: A deep learning based smartphone navigation assistant for people with visual impairments},
journal = {Expert Systems with Applications},
volume = {212},
pages = {118720},
year = {2023},
issn = {0957-4174},
doi = {https://doi.org/10.1016/j.eswa.2022.118720},
url = {https://www.sciencedirect.com/science/article/pii/S0957417422017432},
author = {Bineeth Kuriakose and Raju Shrestha and Frode Eika Sandnes}
}

@article{4,
  title={V-eye: A vision-based navigation system for the visually impaired},
  author={Duh, Ping-Jung and Sung, Yu-Cheng and Chiang, Liang-Yu Fan and Chang, Yung-Ju and Chen, Kuan-Wen},
  journal={IEEE Transactions on Multimedia},
  volume={23},
  pages={1567--1580},
  year={2020},
  publisher={IEEE}
}

@inproceedings{5,
  title={Deep learning based wearable assistive system for visually impaired people},
  author={Lin, Yimin and Wang, Kai and Yi, Wanxin and Lian, Shiguo},
  booktitle={Proceedings of the IEEE/CVF international conference on computer vision workshops},
  pages={0--0},
  year={2019}
}

@article{6,
  title={Gpt-4o system card},
  author={Hurst, Aaron and Lerer, Adam and Goucher, Adam P and Perelman, Adam and Ramesh, Aditya and Clark, Aidan and Ostrow, AJ and Welihinda, Akila and Hayes, Alan and Radford, Alec and others},
  journal={arXiv preprint arXiv:2410.21276},
  year={2024}
}

@article{7,
  title={Visual instruction tuning},
  author={Liu, Haotian and Li, Chunyuan and Wu, Qingyang and Lee, Yong Jae},
  journal={Advances in neural information processing systems},
  volume={36},
  year={2024}
}

@inproceedings{8,
  title={Vqa: Visual question answering},
  author={Antol, Stanislaw and Agrawal, Aishwarya and Lu, Jiasen and Mitchell, Margaret and Batra, Dhruv and Zitnick, C Lawrence and Parikh, Devi},
  booktitle={Proceedings of the IEEE international conference on computer vision},
  pages={2425--2433},
  year={2015}
}

@article{9,
  title={A multi-world approach to question answering about real-world scenes based on uncertain input},
  author={Malinowski, Mateusz and Fritz, Mario},
  journal={Advances in neural information processing systems},
  volume={27},
  year={2014}
}

@inproceedings{10,
  title={Making the v in vqa matter: Elevating the role of image understanding in visual question answering},
  author={Goyal, Yash and Khot, Tejas and Summers-Stay, Douglas and Batra, Dhruv and Parikh, Devi},
  booktitle={Proceedings of the IEEE conference on computer vision and pattern recognition},
  pages={6904--6913},
  year={2017}
}

@article{11,
  title={Are you talking to a machine? dataset and methods for multilingual image question},
  author={Gao, Haoyuan and Mao, Junhua and Zhou, Jie and Huang, Zhiheng and Wang, Lei and Xu, Wei},
  journal={Advances in neural information processing systems},
  volume={28},
  year={2015}
}

@inproceedings{12,
  title={Clevr: A diagnostic dataset for compositional language and elementary visual reasoning},
  author={Johnson, Justin and Hariharan, Bharath and Van Der Maaten, Laurens and Fei-Fei, Li and Lawrence Zitnick, C and Girshick, Ross},
  booktitle={Proceedings of the IEEE conference on computer vision and pattern recognition},
  pages={2901--2910},
  year={2017}
}

@inproceedings{13,
  title={Vizwiz: nearly real-time answers to visual questions},
  author={Bigham, Jeffrey P and Jayant, Chandrika and Ji, Hanjie and Little, Greg and Miller, Andrew and Miller, Robert C and Miller, Robin and Tatarowicz, Aubrey and White, Brandyn and White, Samual and others},
  booktitle={Proceedings of the 23nd annual ACM symposium on User interface software and technology},
  pages={333--342},
  year={2010}
}

@misc{14,
	author = {},
	title = {{B}e {M}y {E}yes - {S}ee the world together --- bemyeyes.com},
	howpublished = {\url{https://www.bemyeyes.com/}},
	year = {},
	note = {[Accessed 14-01-2025]},
}

@article{15,
  title={Gpt-4 technical report},
  author={Achiam, Josh and Adler, Steven and Agarwal, Sandhini and Ahmad, Lama and Akkaya, Ilge and Aleman, Florencia Leoni and Almeida, Diogo and Altenschmidt, Janko and Altman, Sam and Anadkat, Shyamal and others},
  journal={arXiv preprint arXiv:2303.08774},
  year={2023}
}

@inproceedings{16,
  title={VQAsk: a multimodal Android GPT-based application to help blind users visualize pictures},
  author={De Marsico, Maria and Giacanelli, Chiara and Manganaro, Clizia Giorgia and Palma, Alessio and Santoro, Davide},
  booktitle={Proceedings of the 2024 International Conference on Advanced Visual Interfaces},
  pages={1--5},
  year={2024}
}

@misc{17,
	author = {},
	title = {mousi --- mousi.org},
	howpublished = {\url{http://mousi.org/}},
	year = {},
	note = {[Accessed 14-01-2025]},
}

@article{18,
  title={Vialm: A survey and benchmark of visually impaired assistance with large models},
  author={Zhao, Yi and Zhang, Yilin and Xiang, Rong and Li, Jing and Li, Hillming},
  journal={arXiv preprint arXiv:2402.01735},
  year={2024}
}

@article{19,
  title={VIAssist: Adapting Multi-modal Large Language Models for Users with Visual Impairments},
  author={Yang, Bufang and He, Lixing and Liu, Kaiwei and Yan, Zhenyu},
  journal={arXiv preprint arXiv:2404.02508},
  year={2024}
}

@article{20,
  title={Long-Form Answers to Visual Questions from Blind and Low Vision People},
  author={Huh, Mina and Xu, Fangyuan and Peng, Yi-Hao and Chen, Chongyan and Murugu, Hansika and Gurari, Danna and Choi, Eunsol and Pavel, Amy},
  journal={arXiv preprint arXiv:2408.06303},
  year={2024}
}

@inproceedings{22,
  title={Open scene understanding: Grounded situation recognition meets segment anything for helping people with visual impairments},
  author={Liu, Ruiping and Zhang, Jiaming and Peng, Kunyu and Zheng, Junwei and Cao, Ke and Chen, Yufan and Yang, Kailun and Stiefelhagen, Rainer},
  booktitle={Proceedings of the IEEE/CVF International Conference on Computer Vision},
  pages={1857--1867},
  year={2023}
}

@inproceedings{23,
  title={Segment anything},
  author={Kirillov, Alexander and Mintun, Eric and Ravi, Nikhila and Mao, Hanzi and Rolland, Chloe and Gustafson, Laura and Xiao, Tete and Whitehead, Spencer and Berg, Alexander C and Lo, Wan-Yen and others},
  booktitle={Proceedings of the IEEE/CVF International Conference on Computer Vision},
  pages={4015--4026},
  year={2023}
}

@article{24,
  author = {David Burmedi and Stefanie Becker and Vera Heyl and Hans-Werner Wahl and Ines Himmelsbach},
  title = {Emotional and social consequences of age-related low vision},
  journal = {Visual Impairment Research},
  volume = {4},
  number = {1},
  pages = {47--71},
  year = {2002},
  publisher = {Taylor \& Francis},
  doi = {10.1076/vimr.4.1.47.15634},
  url = {https://doi.org/10.1076/vimr.4.1.47.15634},
  eprint = {https://doi.org/10.1076/vimr.4.1.47.15634}
}

@article{25,
  title={Loneliness, adaptation to vision impairment, social support and depression among visually impaired elderly},
  journal={International Congress Series},
  volume={1282},
  pages={317--321},
  year={2005},
  note={Vision 2005},
  issn={0531-5131},
  doi={https://doi.org/10.1016/j.ics.2005.04.017},
  url={https://www.sciencedirect.com/science/article/pii/S0531513105007375},
  author={P.F.J. Verstraten and W.L.J.H. Brinkmann and N.L. Stevens and J.S.A.G. Schouten}
}

@article{26,
  title={Emotionqueen: A benchmark for evaluating empathy of large language models},
  author={Chen, Yuyan and Wang, Hao and Yan, Songzhou and Liu, Sijia and Li, Yueze and Zhao, Yi and Xiao, Yanghua},
  journal={arXiv preprint arXiv:2409.13359},
  year={2024}
}

@article{27,
  title={Emphi: Generating empathetic responses with human-like intents},
  author={Chen, Mao Yan and Li, Siheng and Yang, Yujiu},
  journal={arXiv preprint arXiv:2204.12191},
  year={2022}
}

@article{28,
  title={Improving multi-turn emotional support dialogue generation with lookahead strategy planning},
  author={Cheng, Yi and Liu, Wenge and Li, Wenjie and Wang, Jiashuo and Zhao, Ruihui and Liu, Bang and Liang, Xiaodan and Zheng, Yefeng},
  journal={arXiv preprint arXiv:2210.04242},
  year={2022}
}

@article{29,
  title={Exploring the role of an emotional support and counselling service for people with visual impairments},
  author={Hodge, Suzanne and Barr, Wally and Bowen, Louise and Leeven, Martina and Knox, Paul},
  journal={British Journal of Visual Impairment},
  volume={31},
  number={1},
  pages={5--19},
  year={2013},
  publisher={SAGE Publications Sage UK: London, England}
}

@article{30,
  title={CoMAE: A multi-factor hierarchical framework for empathetic response generation},
  author={Zheng, Chujie and Liu, Yong and Chen, Wei and Leng, Yongcai and Huang, Minlie},
  journal={arXiv preprint arXiv:2105.08316},
  year={2021}
}

@article{31,
  title={ESC-Eval: Evaluating Emotion Support Conversations in Large Language Models},
  author={Zhao, Haiquan and Li, Lingyu and Chen, Shisong and Kong, Shuqi and Wang, Jiaan and Huang, Kexin and Gu, Tianle and Wang, Yixu and Jian, Wang and Liang, Dandan and others},
  journal={arXiv preprint arXiv:2406.14952},
  year={2024}
}

@article{32,
  title={Towards empathetic open-domain conversation models: A new benchmark and dataset},
  author={Rashkin, Hannah},
  journal={arXiv preprint arXiv:1811.00207},
  year={2018}
}

@article{33,
  title={LoRA: Low-Rank Adaptation of Large Language Models},
  author={J. Edward Hu and Yelong Shen and Phillip Wallis and Zeyuan Allen-Zhu and Yuanzhi Li and Shean Wang and Weizhu Chen},
  journal={ArXiv},
  year={2021},
  volume={abs/2106.09685},
  url={https://api.semanticscholar.org/CorpusID:235458009}
}

@article{34,
  title={Direct preference optimization: Your language model is secretly a reward model},
  author={Rafailov, Rafael and Sharma, Archit and Mitchell, Eric and Manning, Christopher D and Ermon, Stefano and Finn, Chelsea},
  journal={Advances in Neural Information Processing Systems},
  volume={36},
  year={2024}
}

@inproceedings{35,
  title={Improving automatic vqa evaluation using large language models},
  author={Ma{\~n}as, Oscar and Krojer, Benno and Agrawal, Aishwarya},
  booktitle={Proceedings of the AAAI Conference on Artificial Intelligence},
  volume={38},
  number={5},
  pages={4171--4179},
  year={2024}
}

@inproceedings{36,
    title = "{ROUGE}: A Package for Automatic Evaluation of Summaries",
    author = "Lin, Chin-Yew",
    booktitle = "Text Summarization Branches Out",
    month = jul,
    year = "2004",
    address = "Barcelona, Spain",
    publisher = "Association for Computational Linguistics",
    url = "https://aclanthology.org/W04-1013/",
    pages = "74--81"
}

@inproceedings{37,
    title = "{METEOR}: An Automatic Metric for {MT} Evaluation with High Levels of Correlation with Human Judgments",
    author = "Lavie, Alon  and
      Agarwal, Abhaya",
    editor = "Callison-Burch, Chris  and
      Koehn, Philipp  and
      Fordyce, Cameron Shaw  and
      Monz, Christof",
    booktitle = "Proceedings of the Second Workshop on Statistical Machine Translation",
    month = jun,
    year = "2007",
    address = "Prague, Czech Republic",
    publisher = "Association for Computational Linguistics",
    url = "https://aclanthology.org/W07-0734/",
    pages = "228--231"
}

@article{38,
  title={Bertscore: Evaluating text generation with bert},
  author={Zhang, Tianyi and Kishore, Varsha and Wu, Felix and Weinberger, Kilian Q and Artzi, Yoav},
  journal={arXiv preprint arXiv:1904.09675},
  year={2019}
}

@article{40,
  title={Emotional intelligence of large language models},
  author={Wang, Xuena and Li, Xueting and Yin, Zi and Wu, Yue and Liu, Jia},
  journal={Journal of Pacific Rim Psychology},
  volume={17},
  pages={18344909231213958},
  year={2023},
  publisher={SAGE Publications Sage UK: London, England}
}

@article{41,
  title={EmoBench: Evaluating the Emotional Intelligence of Large Language Models},
  author={Sabour, Sahand and Liu, Siyang and Zhang, Zheyuan and Liu, June M and Zhou, Jinfeng and Sunaryo, Alvionna S and Li, Juanzi and Lee, Tatia and Mihalcea, Rada and Huang, Minlie},
  journal={arXiv preprint arXiv:2402.12071},
  year={2024}
}

@article{42,
  title={Expanding performance boundaries of open-source multimodal models with model, data, and test-time scaling},
  author={Chen, Zhe and Wang, Weiyun and Cao, Yue and Liu, Yangzhou and Gao, Zhangwei and Cui, Erfei and Zhu, Jinguo and Ye, Shenglong and Tian, Hao and Liu, Zhaoyang and others},
  journal={arXiv preprint arXiv:2412.05271},
  year={2024}
}

@article{43,
  title={Qwen2-vl: Enhancing vision-language model's perception of the world at any resolution},
  author={Wang, Peng and Bai, Shuai and Tan, Sinan and Wang, Shijie and Fan, Zhihao and Bai, Jinze and Chen, Keqin and Liu, Xuejing and Wang, Jialin and Ge, Wenbin and others},
  journal={arXiv preprint arXiv:2409.12191},
  year={2024}
}

\end{document}